\documentclass[10pt,twocolumn,letterpaper]{article}

\usepackage{iccv}
\usepackage{times}
\usepackage{epsfig}
\usepackage{graphicx}
\usepackage{amsmath}
\usepackage{amssymb}

\usepackage{multirow}

\usepackage[pagebackref=true,breaklinks=true,letterpaper=true,colorlinks,bookmarks=false]{hyperref}
\iccvfinalcopy % *** Uncomment this line for the final submission

\ificcvfinal\fi

\begin{document}

%%%%%%%%% TITLE
\title{Rethinking Annotation for Object Detection:\\ Is Annotating Small-size Instances Worth Its Cost?}

\author{%
  Yusuke Hosoya${}^\text{1}$
\quad
Masanori Suganuma${}^\text{1}$
\quad
Takayuki Okatani${}^\text{1,2}$\\
${}^\text{1}$Graduate School of Information Sciences, Tohoku University
\quad
${}^\text{2}$RIKEN Center for AIP\\
{\tt\small \{yhosoya,suganuma,okatani\}@vision.is.tohoku.ac.jp}
}

\maketitle
\ificcvfinal\thispagestyle{empty}\fi

%%%%%%%%% ABSTRACT
\begin{abstract}
Detecting objects occupying only small areas in an image is difficult, even for humans. Therefore, annotating small-size object instances is hard and thus costly. This study questions common sense by asking the following: is annotating small-size instances worth its cost? We restate it as the following verifiable question: can we detect small-size instances with a detector trained using training data free of small-size instances? We evaluate a method that upscales input images at test time and a method that downscales images at training time. The experiments conducted using the COCO dataset show the following. The first method, together with a remedy to narrow the domain gap between training and test inputs, achieves at least comparable performance to the baseline detector trained using complete training data. Although the method needs to apply the same detector twice to an input image with different scaling, we show that its distillation yields a single-path detector that performs equally well to the same baseline detector. These results point to the necessity of rethinking the annotation of training data for object detection. 
\end{abstract}

%%%%%%%%% BODY TEXT
\section{Introduction}

Object detection has been extensively studied so far. The application of convolutional neural networks (CNNs) has brought about significant performance improvements. The mainstream approach is to train CNNs or other deep networks in a supervised fashion using a large dataset manually annotated by humans. Several datasets have been created, such as PascalVOC \cite{VOC}, COCO \cite{MSCOCO}, Open Images Dataset \cite{OpenImages}, and KITTI \cite{KITTI}, which contributed to the advancement of studies. 

\begin{figure}[tb]
\centering
\includegraphics[width=0.95\linewidth,bb=0 0 2907 1536]{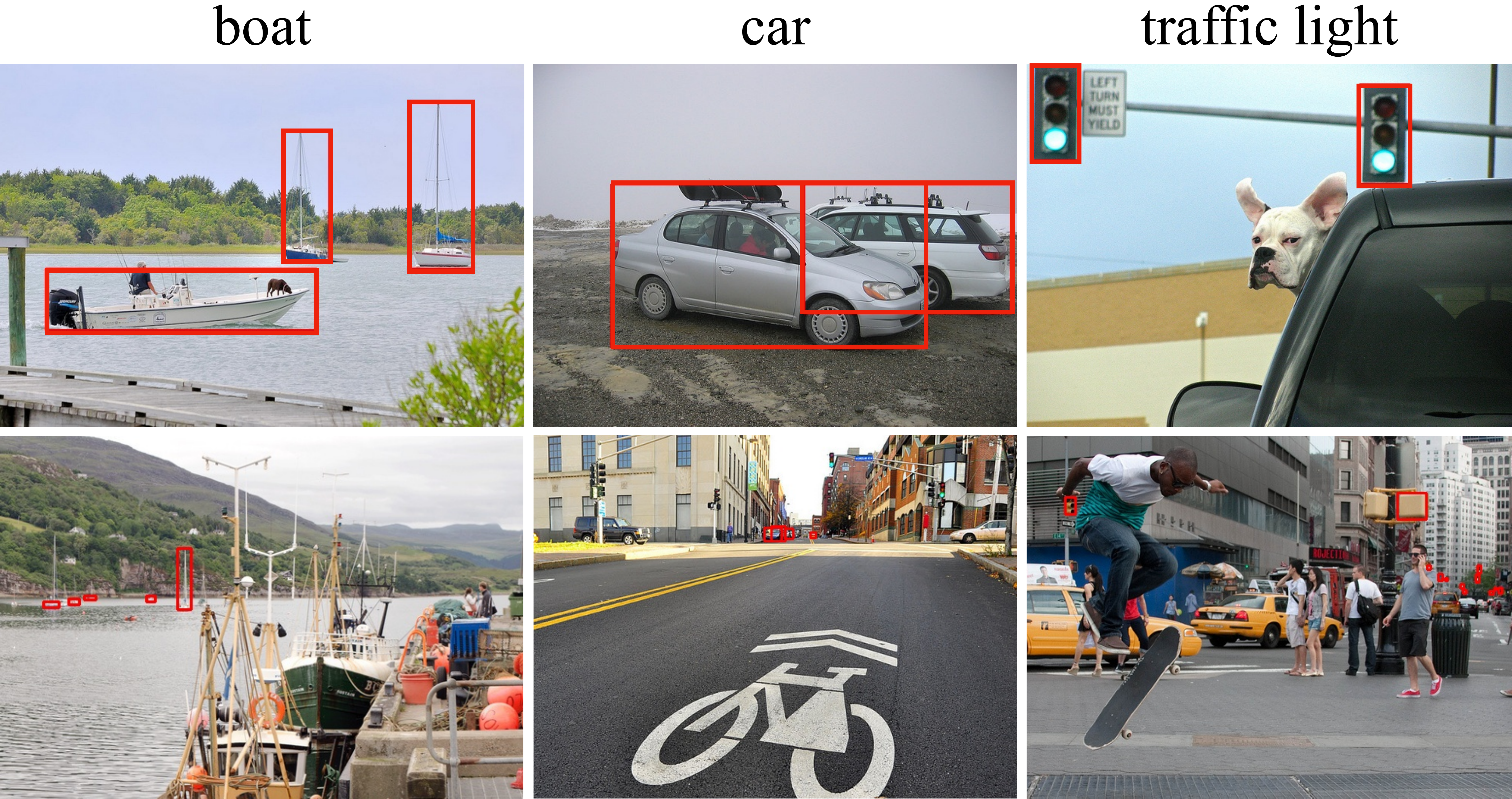}
\caption{
Compared with mid- and large-size object instances (upper row), small-size instances are harder to annotate correctly (lower row).
}
\label{fig:small_example}
\end{figure}

In this paper, we reconsider the image annotation for object detection. It is a procedure of providing a bounding box and a class label to each object instance in images in some ways \cite{ExtremeClick, LearnDialog, ClickSV}. Since it tends to be costly, it is essential to know {\em how to achieve the most significant effect with the minimum cost} (i.e., the model trained on the annotated data achieves a higher performance). This thinking is crucial from the practitioner's perspective, although researchers have not paid much attention. 

In this paper, we shed light on the annotation of objects occupying small areas in images. In its background, there is a fact that annotating such small instances tends to be more costly than medium or large-size instances. To be specific, since they are small, annotators tend to fail to find them, or it is hard to make an error-free judgment about their class or even existence; see Fig.~\ref{fig:small_example}. Besides, small-size instances tend to appear as a large population (e.g., a crowd of people), making it cumbersome to label them one by one, as shown in Fig.~\ref{fig:COCO_errlabels}. Thus, annotators need more time for annotating small-size instances. The smaller the size is, the more pronounced these tendencies will be.

Having these in mind, we pose the following question: {\em is the annotation of small object instances worth its cost? } Instead of attempting to answer it directly, we consider if {\em we can train a detector to detect small-size instances without annotating them.}
Given the basic fact that an object's apparent size is determined relative to its distance, we consider resizing input images either at training time or at test time to see if we can do the above. 

An obstacle against this approach is the potential domain gap emerging between training data and test data. Synthetic small-size instances created by downscaling non-small-size instances will have a different appearance from actual small-size instances. Upscaled object instances that are small in their native size will have a different appearance from actual non-small-size instances. There are two possible types of appearance differences. One is the difference in image resolution; upscaling small-size instances makes their appearance blurry. The other is the difference in the remaining factors, such as an object's perspective, e.g., determined by the focal length of the camera used to capture the image, and context, such as its surrounding area in the image.

To investigate the (un)importance of small-size instance annotation while having the above obstacle in mind, we consider two methods and evaluate their performance. One is the method that upscales images when inputting them to a detector at test time, aiming to make it detect small-size instances. It trains a detector using only object instances other than small-size ones. The other is the method that downscales images at training time. It downscales non-small-size object instances and uses them for training a detector. We use the COCO dataset since it is the most widely used for object detection in the community. We employ its classification into small/medium/large-size instances. 

We experimentally evaluate the above two methods, finding an answer to the above question about the importance of small-size instance annotation. The answer is that we can detect small-size object instances with a detector trained on a dataset free of small-size instances. Specifically, the method of upscaling images at test time, together with a remedy to narrow the domain gap between training and test inputs, achieves at least comparable performance to the baseline detector trained using complete training data. The experimental results also show that the method of downscaling images at training time does not work well, which is arguably attributable to the above domain gap. Additionally, although the upscaling method needs to apply the same detector twice to an input image with different scaling to detect all instance sizes, we show that its distillation yields a single-path detector that shows comparable performance to the baseline detector. These results point to the necessity of rethinking the annotation of training data for object detection.

\section{Related Work}

\subsection{Object Detectors}

Many studies have been conducted on object detection, and we have witnessed dramatic performance improvement with the recent advancements of convolutional neural networks in their architectural design and training method.
Existing object detectors are classified into anchor-based methods, anchor-free methods, and others. The anchor-based methods are further classified into one-stage methods, including SSD \cite{SSD}, YOLOv2 \cite{yolov2} and many others \cite{DSSD,yolov3,RefineDet,RetinaNet,M2Det} and two-stage methods, such as Faster RCNN \cite{FasterRCNN} and others \cite{R_FCN,MaskRCNN,CascadeRCNN}. These methods use default bounding boxes (BBs) with unique positions and shapes, called anchor boxes, and detect objects by matching these default boxes to them. The anchor-free methods do not use anchor boxes. Some methods \cite{CornerNet,CenterNet,ExtremeNet,ObjectsAsPoints,GridRCNN,RepPoints} formulate the problem as detecting keypoints and others \cite{FCOS,FSFA,FoveaBox} predict offsets to the edges of object BBs from grid points defined on feature maps. 

As discussed earlier, it is hard or almost impossible to create an error-free dataset for object detection. 
Although annotation errors will impact both training of detectors and their evaluation, only a few studies estimate how large the impact will be or how to cope with them, apart from the studies dealing with label noise for image classification tasks \cite{ProbNoiseCorrection,JointOptim,LearnReweight,mentornet,Decoupling,coteaching} to name a few. Li \etal \cite{NR_OD} point out that there are two types of annotation errors, those in object class labels and those in BB specification. They then propose a method alternately repeating two steps, using the current detector to correct erroneous BBs and class labels and updating the detector. Samet \etal \cite{ReduceNoise} propose a method for better training anchor-free detectors in the presence of errors in BBs' geometry. It trains a detector with a mechanism of adjusting BBs' geometry while down-weighting the regions of objects' background and occlusion that contain irrelevant image features and thus could harm the detector's training.

\subsection{Using Multi-scale Images}
Objects have various apparent sizes on images. It has been a primary concern in the research of object detection to deal with them properly. Many studies have been conducted, particularly on how to accurately detect small-size object instances. For instance, some methods use multi-scale feature maps \cite{SSD,DSSD,RefineDet} or a feature pyramid \cite{FPN}, and others aggregate them into a single feature map \cite{M2Det,PANet,EFDet}.

A more straightforward approach is multi-scale test. An input image is resized to form multi-scale images, which are inputted to a detector, yielding BBs at the multiple scales. They are merged by non-maximum suppression etc. to produce the final result. Multi-scale training does this at training time, i.e., to use multi-scale images obtained by resizing an input image for training. SNIP (Scale Normalization for Image Pyramids) \cite{SNIP} improves multi-scale training by choosing only instances of proper sizes for each scale (e.g., small and medium instances in enlarged images or medium and large instances in shrunken images). Extending this, SNIPER (SNIP with Efficient Resampling) \cite{SNIPER} shows that it is not always good to input high-resolution images for training a detector and proposes to use sub-images properly cropped from them for training, enabling memory-efficient and fast multi-scale training.

There is an argument \cite{GS_Face} that the low accuracy of the detection of small-size instances is because of the imbalance between the number of positive and negative samples assigned to anchors with them. It shows that by mitigating the imbalance by the proposed sampling method, face detection can be performed as accurately by FPN \cite{FPN} with only the feature map of the highest resolution
%at the highest-scale (i.e., the lowest resolution{\color{cyan}[FPNの最終層なのでhighest resolution?]}) 
as by FPN with the full multi-scale feature maps.

It is also reported \cite{FixResolution} for object recognition that the standard data augmentation causes a discrepancy in apparent object sizes between training and test, harming classification accuracy.
The study shows that minimizing the gap by making training images smaller or test images larger can rectify this.

\subsection{Semi-/Weakly-supervised Learning}

Another approach is to formulate object detection in the framework of semi-supervised or weakly-supervised learning, which aims mainly at reducing annotation cost but will also contribute to mitigate or eliminate the effects of erroneous annotation. There are several studies on transferring knowledge from image-level labels to object detection
\cite{LS_SSOD,RevisitKnowTrans,AppearanceTrans}; consistency constraints are imposed on classification and regression in \cite{ConsistencySSOD} . 
Gao \etal \cite{NoteRCNN} proposed a method preventing detectors from overfitting to labelling noises emerging in the semi-supervised learning setting, where knowledge distillation is employed using a pretrained detector with clean annotation as a teacher. Although they are attractive, these studies based on semi- or weakly-supervised learning including the recent ones \cite{LatentV,ObjectAware} achieves only inferior performance to the method based on supervised learning utilizing fully labeled training data.

\subsection{Methods for Annotation and Evaluation}%\subsection{Dataset Creation}

Creating datasets for object detection, especially annotating object BBs, is labor-intensive.  Several methods have been proposed that make annotation of BBs easier, such as clicking only four points to specify an object's BB \cite{ExtremeClick}, which is employed for annotating BBs in the Open Images dataset \cite{OpenImages}, clicking only an object's center \cite{ClickSV}, and choosing from the candidates provided \cite{LearnDialog}. Some datasets are automatically generated from existing data of semantic/instance segmentation, e.g., the COCO dataset. 

To evaluate the accuracy of object detection, average precision (AP), which approximates the area under the recall-precision curve, has been used as the standard metric \cite{VOC,MSCOCO}. A few studies question its optimality as an evaluation metric. There is an argument \cite{LRP_Det} that AP cannot represent fine differences of detection accuracy, such as the difference between high precision/low recall and low precision/high recall, leading to the proposal of a new metric. It is pointed out \cite{AverageDelay} that AP does not capture temporal delay that emerges in video object detection, and a new metric measuring this is proposed. 

\section{Annotating Small-size Instances}

\subsection{Annotation Difficulties}

As is shown in Fig.~\ref{fig:small_example}, it is generally hard to annotate small-size object instances. It is hard because the smaller the object instances are, the more closer recognizing them is to the upper limit of the human visual ability. First, annotators are more likely to overlook small-size instances because of their sizes. Annotators need to search everywhere in images. Besides, for small-size visual entities, it is often hard for annotators to judge if they are objects to annotate or what class they belong to if they are. This will inevitably increase in annotation time and cost. 

\begin{figure*}[t]
\centering
\includegraphics[width=1.0\linewidth,bb=0 0 5441 1806]{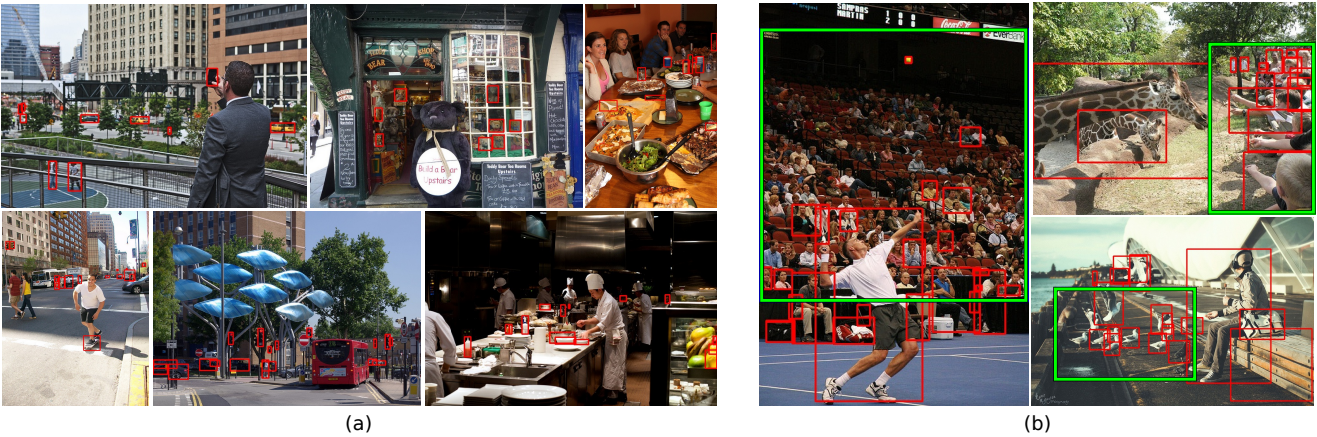}
\caption{
Examples showing difficulties with annotating small-size object instances. (a) Images that are hard to annotate; while careful inspection reveals several BBs are missing, some are really hard to judge. Only small-size instances are visualized. (b) In the COCO dataset, annotators can specify bounding boxes with a binary label ``iscrowd'' (shown in green) to enclose a crowd of objects; they are excluded from both training and evaluation.
}
\label{fig:COCO_errlabels}
\end{figure*}

The annotation is cumbersome when there are many instances in an image, as shown in Fig.~\ref{fig:COCO_errlabels}. Such a case most often occurs for the object class of persons (i.e., a crowd of people), merchandise displayed in a storefront, and ingredients in a dish. To avoid an increase in annotation costs, COCO employs a rule setting the upper bound on the number of instances to annotate in an image per object class. It provides a special label ``$iscrowd$'' for the purpose.  Annotators need to annotate ten instances or fewer per object/image\footnote{Some images have fifteen object instances.}. They are allowed to specify a large bounding box enclosing small-size instances they choose not to individually annotate while setting $iscrowd=yes$ with it, as shown in Fig.~\ref{fig:COCO_errlabels}(b). The bounding boxes with this label are not used at training time and excluded when evaluating detectors' performance, avoiding the above issue. However, it is left to the annotators' judgment about when to use this option, and the above difficulty remains unsolved.

\subsection{Detection Accuracy for Small-size Instances}

Due to its difficulty, annotating small-size instances will be more prone to errors. We conjecture that small-size instance annotation does have more errors, e.g., in the COCO dataset. However, this conjecture is not so easy to validate. It involves many complicated factors, such as the bound of human vision regarding how small objects it can recognize precisely. Thus, we instead performed a simple experiment, whose results provide circumstantial evidence. 

\begin{figure}[t]
\centering
\includegraphics[width=1.0\linewidth,bb=0 0 3000 1664]{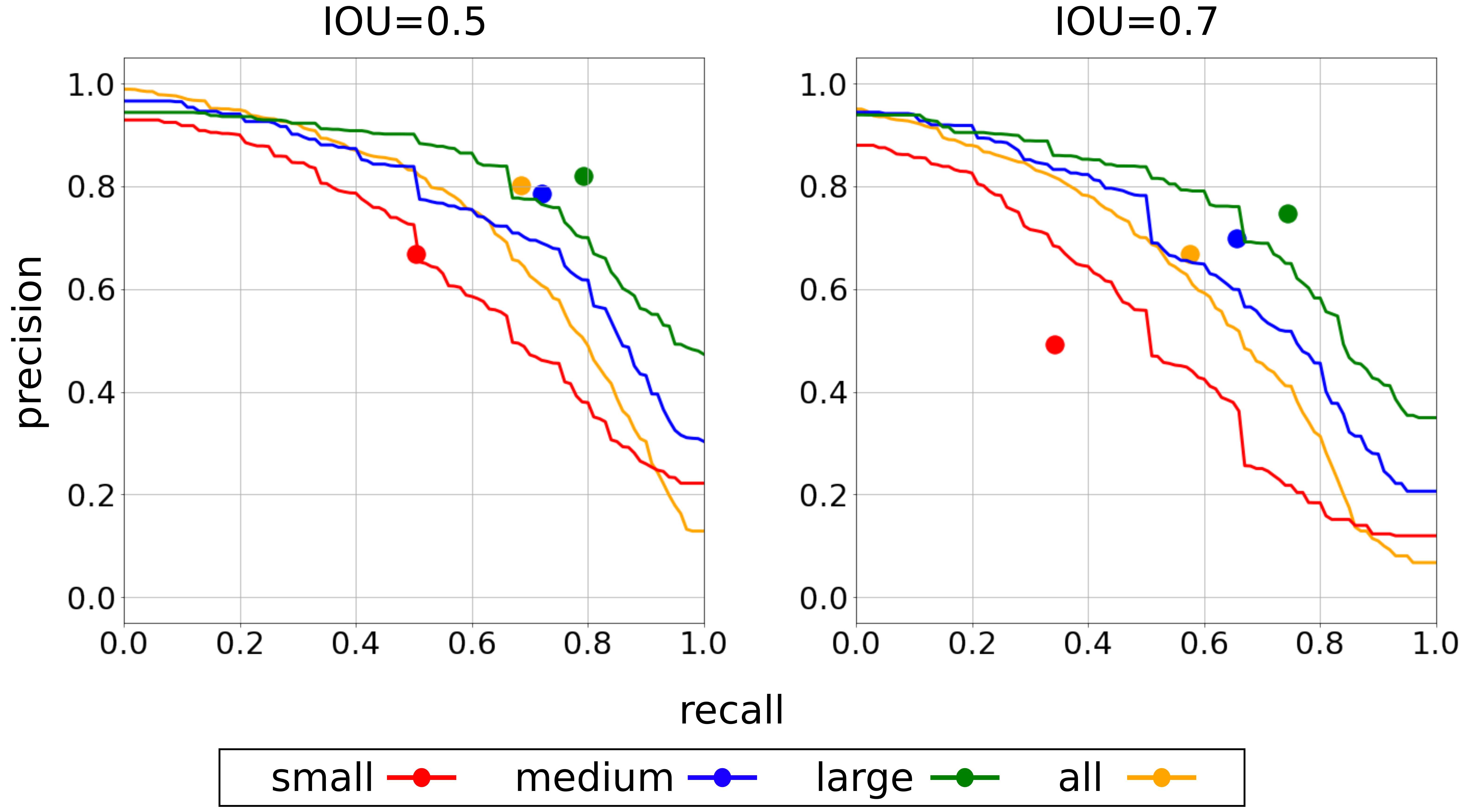}
\caption{
``Accuracy'' of a human subject's annotation on selected 500 images from the COCO validation split, plotted in colored dots. The precision-recall curves of SOTA object detector, EfficientDet-D7 \cite{EFDet}, are shown for comparison. Evaluation with two definitions of correct detection  (IOU threshold with $0.5$ and $0.7$). 
}
\label{fig:cnn_vs_self}
\end{figure}

One of the authors of the present paper annotated 500 images randomly chosen from the validation split of COCO 2017 dataset. Regarding his annotation work as a detector, we can evaluate its ``accuracy'' using the ground-truth annotation provided by COCO. Figure \ref{fig:cnn_vs_self} shows the results. For the sake of comparison, we also show the accuracy of one of the current state-of-the-art detectors, EfficientDet-D7 \cite{EFDet}. The object instances are divided according to the standard procedure of COCO into three categories, {\em small}, {\em medium}, and {\em large}.

Two observations can be made from Fig.~\ref{fig:cnn_vs_self}. One is that his performance for the small-size object instances is very bad and far from the upper-right corner (i.e., precision$=1$ and recall$=1$). It is the outcome of meticulous work, and its accuracy should be close to the upper bound performance of average humans. It is then natural to think the result implies the limit of annotation suggested above. The other observation is that while he is superior by a certain margin to EfficientDet for medium and large instances, his superiority vanishes for small instances. Do SOTA detectors already attain human-level performance for small-size instances, contrary to the widespread recognition that detection accuracy for them needs to be improved?

We do not go into more details in this paper. We instead consider how important small-size instance annotation is for training object detectors in what follows.

\section{Methods}

Our basic idea is to exclude annotation for small object instances and train a detector with only instances having more than a certain size (i.e., medium and large instances in COCO). To make it nevertheless detect small-size objects, we control the size of input images either at training time or test time. As it is separated from detector's algorithms, this can be used with any detector. In our experiments, we use Faster RCNN \cite{FasterRCNN} and FCOS \cite{FCOS} as representatives of anchor-based and anchor-free detectors, respectively.

\subsection{Upscaling Images at Test Time}\label{sec:uptest}
\begin{figure*}[t]
\hfil
\includegraphics[width=1.0\textwidth,bb=0 0 983 255]{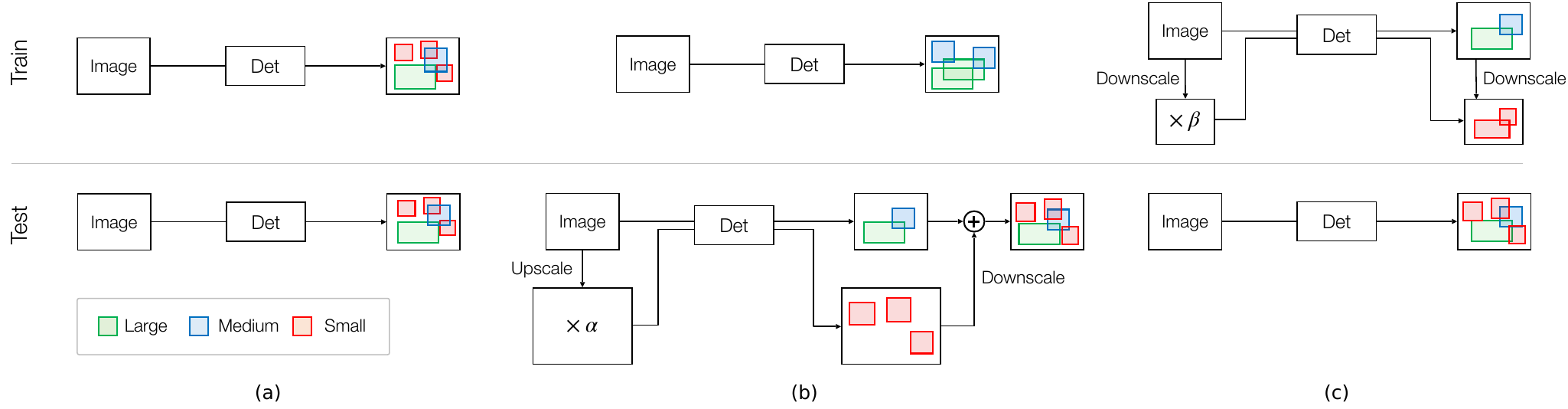}
\caption{Three methods compared in the experiments. (a) The ordinary method. A model is trained with and applied to images with the native resolution for all size instances. (b) {\em Up@Test}: upscaling input images at test time. A model is trained only with medium- and large-size instances. It detects small-size instances from upscaled images and others from the original images. (c) {\em Down@Train}: downscaling input images at training time. A model is trained with images with the native resolution to detect medium and large instances and trained with downscaled images to detect small-size instances. It detects all size instances from input images with the native resolution. }
\label{fig:three_models}
\end{figure*}

Figure \ref{fig:three_models}(b) illustrates this method. We train a detector with the only medium- and large-size instances. As a result, the detector will not be able to detect small-size instances properly. Therefore, we upscale an input image at test time and feed it to the detector. More precisely, we apply the detector twice to the same image. Firstly, an input image is inputted (without resizing) to the detector, which detects medium- and large-size object instances. The same image is then resized with a magnification rate of $\alpha(>1)$ and then inputted to the detector, which detects small-size instances. The detected bounding boxes are resized with the rate of $1/\alpha$ and merged with those detected in the first pass. Note that we can perform the two computations in parallel. We will refer to this method as {\em Up@Test} below.

A possible problem with this method is that the upscaled images of a small-size object instance will be different from their ideal counterparts, i.e., images of medium- or large-size instances with their native resolution. Typically, the upscaled images will be blurry without fine details. The medium- and large-size instances do not have such blurs, which could generate a gap between training and test samples, leading to performance deterioration.

There will be two potential remedies to the problem. One is to apply super-resolution to the input image when upscaling it to match the images of medium- and large-size instances in terms of sharpness. The other is to use blurred images at training time. Specifically, we transform the training images to appear equally blurry to the upscaled images of a small-size instance; we then use them together with the original images. Several methods can be employed to make the images blurry. We test two methods in our experiments. One is a Gaussian blur with the scale $\sigma$ (in pixels). The other is downscale-and-upscale, i.e., downscaling the image by a factor of 
% $1/\alpha$, followed by upscaling by $\alpha$.
$1/\gamma$, followed by upscaling by $\gamma$. Detectors using anchor boxes do no longer need those for small instances, but we do not eliminate them with Faster RCNN in our experiments.

\subsection{Downscaling Images at Training Time}\label{sec:downtrain} % DownTrain

Figure \ref{fig:three_models}(c) illustrates the method. We train the detector using original medium- and large-size instances and also their downscaled versions. Some of the resized instances will work as small-size instances.
Unlike the above method, the resulting detector has learned to detect small-size instances. Thus, we can use it in the same way as the ordinary detectors; that is, we feed an input image without resizing it. We will refer to this method as {\em Down@Train} below.

We employ bilinear interpolation to resize the images and test a scaling ratio of $\beta=1/2$, $1/3$, and $1/4$ in our experiments. We choose these, as the small-size and medium-size instances in the COCO dataset have the areas in the range of  $[0,32^2)$ and $[32^2,96^2)$ pixels, respectively; thus, the medium-size instances resized by the factor of $1/3$ have the same range as the small-size instances. 

There will be little or no gap in their image resolution, as downscaling an image gives effects similar to image acquisition using a remote camera.

\subsection{Domain Gap between Train and Test Inputs}

The above two methods differ in that one upscales input small-instances at test time to match medium- or large-size instances in the training set whereas the other downscales them to match small-instances we encounter at test time. However, they are the same in that there will emerge a gap between the training and test samples in a statistical sense. 

There are two types of gaps. One is caused by the difference in resolution. As mentioned above, this occurs only for the first method of upscaling inputs at test time. We expect this gap can be made small or negligible by simple data augmentation. The other emerges due to other causes, such as differences in the context and perspective of objects. It is less likely that medium-size or larger instances form an apparent crowd within an image, which leads to a difference in the context, i.e., their surrounding areas. Besides, the downscaled instances will appear as if they were taken by a telephoto lens camera, leading to a difference in perspective. These differences will emerge independently of whether upscaling test images or downscaling training images; the two methods both could suffer from them.

\subsection{Distillation into a Single-path Model}

The first method, {\em Up@Test}, feeds an input image into a detector CNN twice with different scaling. As its computational cost is twice as the base detector, it may not be fair to say {\em Up@Test} is comparable to the base detector if it is so in terms of detection accuracy. Therefore, we consider the distillation of the model into a single-path model. Using the two-path model as a teacher, we generate pseudo labels (annotations) for small object instances and train a single-path student model using them together with the original annotation for med-size and large-size instances. The details are given in Sec.~\ref{sec:exp}.

\section{Experimental Results}
\label{sec:exp}

We conduct experiments to examine what we have discussed so far. 
We use the COCO 2017 dataset; its train split for training and validation split for test. 

\subsection{Experimental Configuration}

We employ Faster RCNN \cite{FasterRCNN} and FCOS \cite{FCOS} as the base detectors. For their backbone networks, we employ ResNet50 \cite{ResNet} and ResNet101-FPN \cite{FPN}, respectively; both are ImageNet pretrained models provided by PyTorch. 
The input image size is $600\times 1000$ and $800\times 1333$, respectively. These are the input sizes at the $\times 1$ scale for training.
We use bilinear interpolation for the downscaling and upscaling operations. We do not employ any kind of multi-scale test or multi-scale training other than ours. 

We evaluate the methods in the standard way, i.e., using all the images with small/medium/large-size instances of the validation set of the COCO 2017 dataset. We use the standard evaluation metric, i.e., the mean average precision over IOU in the range $[0.50,0.95]$. We evaluate the following methods.

\medskip \noindent
{\bf Baseline}~~ The standard use of one of the above two detectors trained in the usual way. Object instances of all sizes are used for training. 
  
\medskip \noindent
{\bf Upscaling at test} ({\em Up@Test})~~ We upscale images at test time, as explained in Sec.~\ref{sec:uptest}, while we do not use the original small-size instances for training. We test multiple upscaling factors $\alpha$'s sampled from the range $[1.0,4.0]$. We consider three variants, which we will refer to as {\em Up@Test-vanilla, -blur, and -scale}. 

\smallskip \noindent
{\em  Up@Test-vanilla}~ The model is simply trained with the original images containing only medium- and large-size instances. 

\smallskip \noindent
{\em Up@Test-blur}~ Gaussian blur is applied to the training images and mixed with the original clean versions by a ratio of $0.414:0.586$, the ratio of the number of small-size instances to the number of medium- and large-size instances in the COCO training set. 
For the Gaussian blur, we test $(k,\sigma)=(7,1.4)$ and $(11,2.0)$, where $k$ is the kernel size and $\sigma$ is the scale of the Gaussian kernel, both in pixels.

\smallskip \noindent
{\em Up@Test-scale}~ The training images are downscaled by a factor $1/\gamma$ and then upscaled by $\gamma$ instead of Gaussian blur. We use $\gamma=2$ and $3$.
The rest is identical.
  
\medskip \noindent
{\bf Downscaling at training} ({\em Down@Train})~~We downscale images at training time, as explained in Sec.~\ref{sec:downtrain}. We use the downscaling factor of $\beta=1/2$, $1/3$, and $1/4$. We do not use the original small-size instances for training. Due to the image downscaling,  the medium- and large-size instances are downscaled accordingly and  serve as small-size instances. We mix the downscaled images and the original images with the ratio of $0.414:0.586$ for training; see the explanation of {\em Up@Test-blur}.

\subsection{Upscaling Images at Test Time}

\begin{figure}[tb]
\includegraphics[width=1.0\linewidth,bb=0 0 3000 1447]{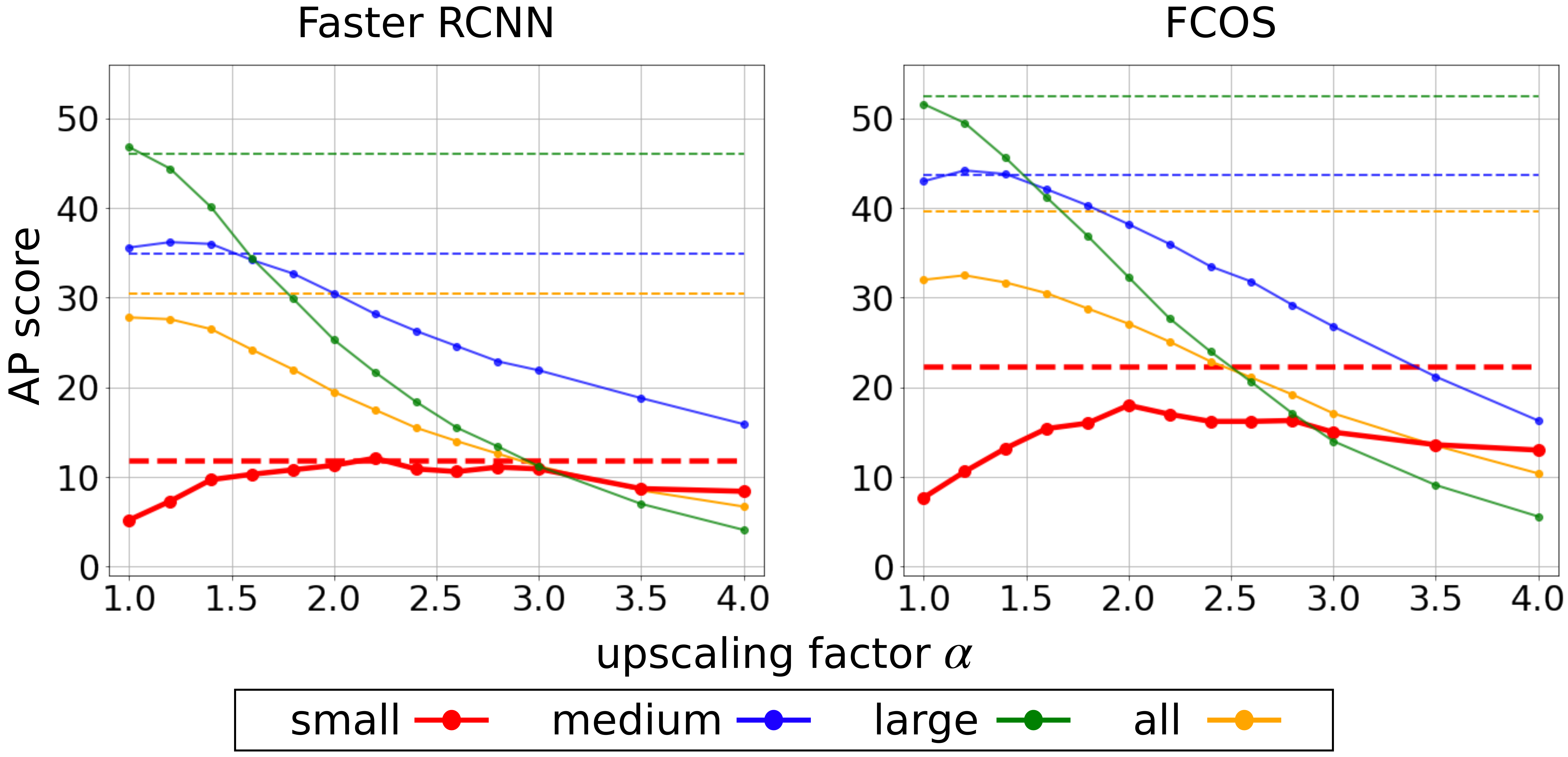}
\caption{
Results of {\em  Up@Test-vanilla}, i.e., trained with the original images and with only medium- and large-size instances. The left and right show Faster RCNN and FCOS, respectively. The horizontal axis indicates the factor of the test-time upscaling. The broken lines indicate {\em Baseline}. 
}
\label{fig:uptest_vanilla}
\end{figure}

Figure \ref{fig:uptest_vanilla} shows the results of {\em  Up@Test-vanilla}. We can make several observations. First, the scores for small-size instances at $\alpha=1.0$ are significantly lower than {\em Baseline} (i.e., the broken lines), unlike those for medium- and large-size instances. This is reasonable as {\em  Up@Test-vanilla} has not learned small-size instances at all. However, the scores increase with $\alpha$ and have a peak at around $\alpha=2.0$ for both detectors. Moreover, the peak score for Faster RCNN matches {\em Baseline}. On the other hand, the scores for medium- and large-size instances monotonically decreases as $\alpha$ increases. This is reasonable, too, since {\em  Up@Test-vanilla} at $\alpha=1.0$ has learned medium- and large-size instances similarly to {\em Baseline} but has not learned them in the original sizes at $\alpha>1$. 

\begin{figure}[t]
\centering
\includegraphics[width=1.0\linewidth,bb=0 0 3000 2621]{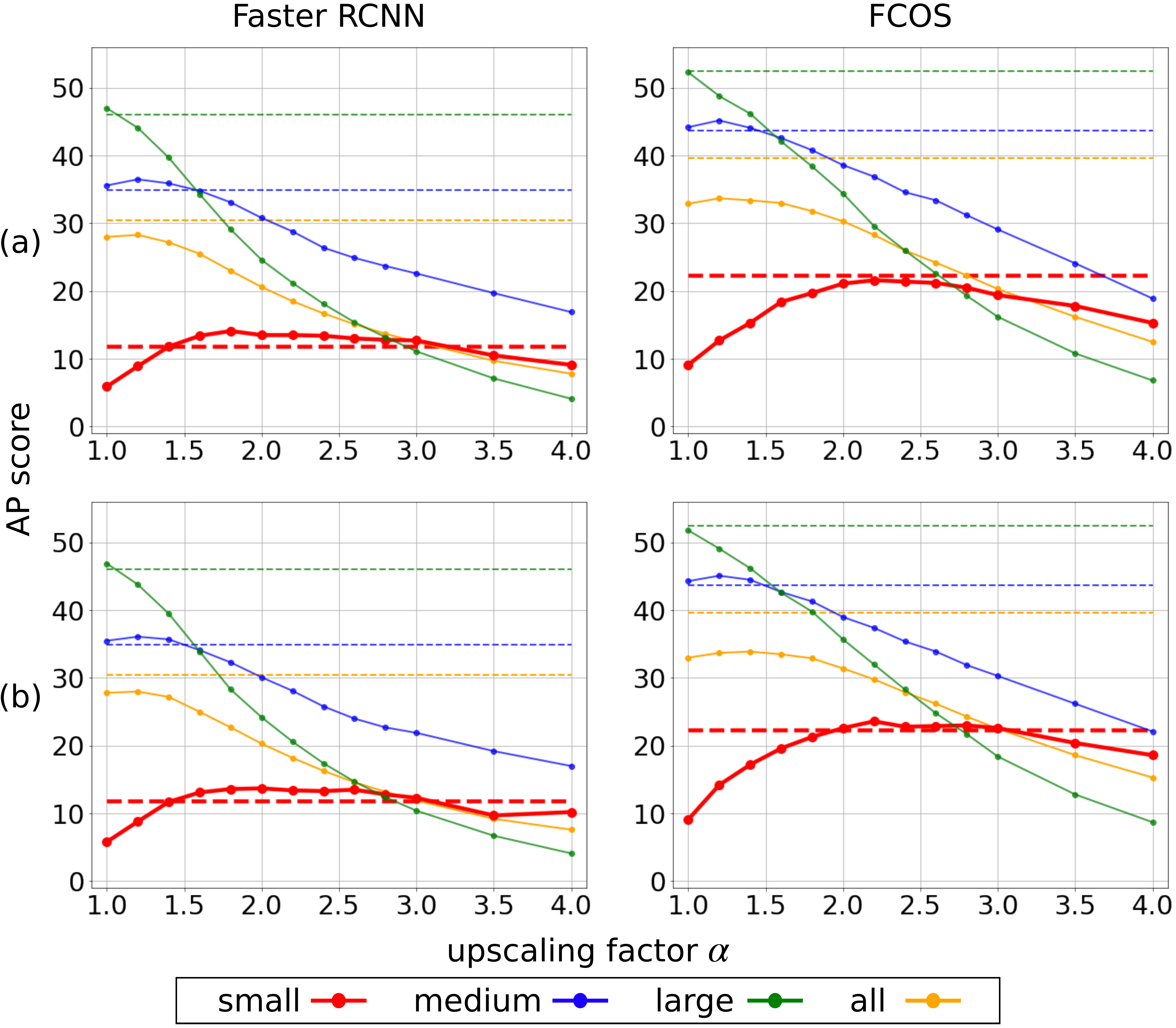}
\caption{Results of {\em Up@Test-scale}. The training images are scaled ($\times 1/\gamma$) and rescaled ($\times \gamma$). (a) $\gamma=2$ and (b) $\gamma=3$. The rest is the same as Fig.~\ref{fig:uptest_vanilla}. }
\label{fig:uptest_scale}
\end{figure}

\begin{figure}[t]
\centering
\includegraphics[width=1.0\columnwidth,bb=0 0 3000 2620]{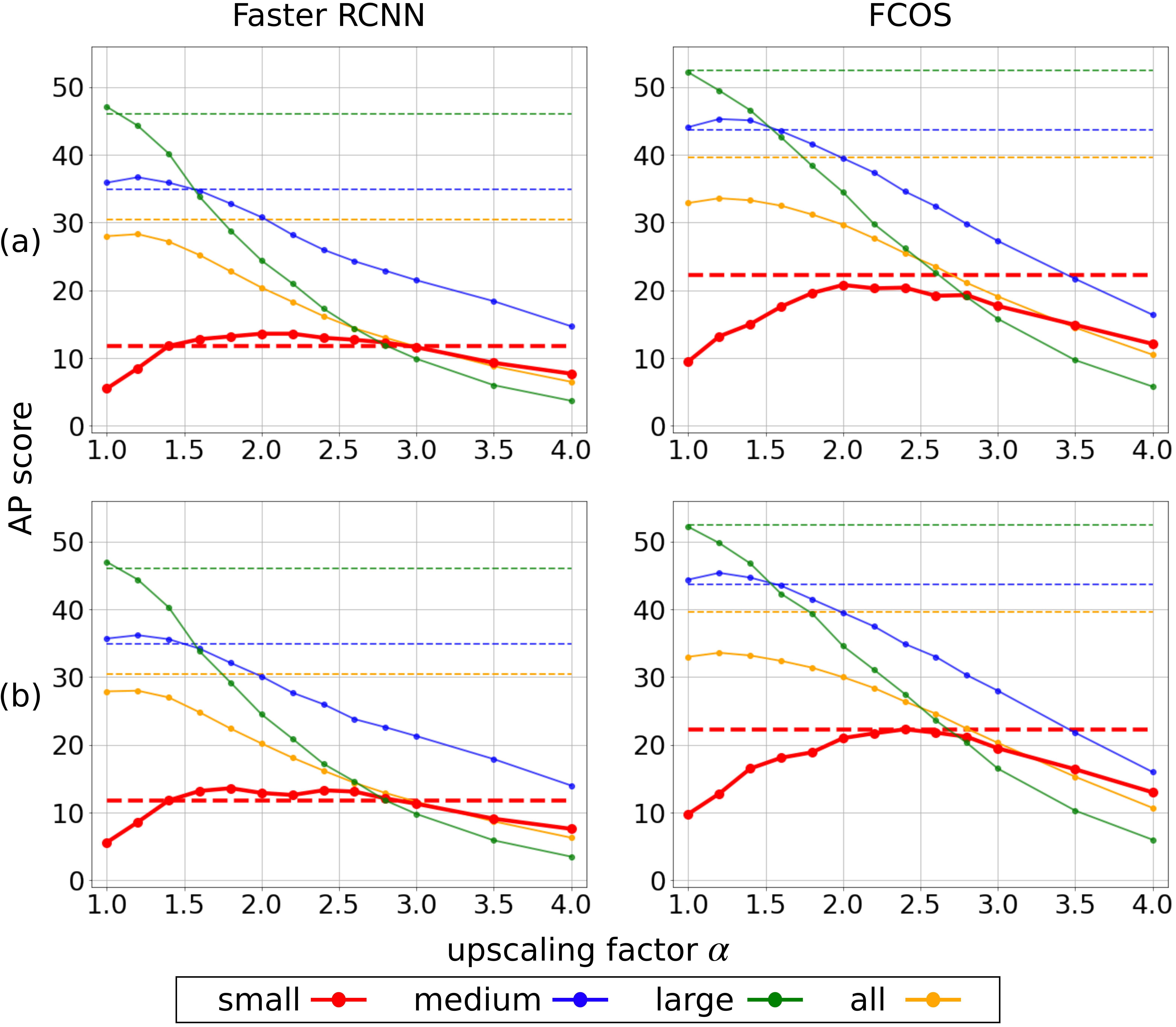}
\caption{
Results of {\em Up@Test-blur}. The training images are blurred with a Gaussian kernel with (a) kernel size $k=7$ and (b) $k=11$. The rest is the same as Fig.~\ref{fig:uptest_vanilla}. 
}
\label{fig:uptest_blur}
\end{figure}

Figures \ref{fig:uptest_scale} and \ref{fig:uptest_blur} show the results of {\em Up@Test-scale} and {\em Up@Test-blur}, respectively. {\em Up@Test-scale} performs the best with $\gamma=3$ and {\em Up@Test-blur} performs the best with $k=11$. Two methods are both effective, but {\em Up@Test-scale} is slightly better. We can observe that it improves the score for small-size objects. As a result, the peak score at around $\alpha=2.0$ matches {\em Baseline} for FCOS and it is even higher for Faster RCNN. The scores for medium- and large-size instances at $\alpha=1.0$ are at least on the same level (or even slightly higher) compared with {\em Baseline} for both detectors, which indicates that they can detect those instances equally well.

\subsection{Downscaling Images at Training Time}

As a detector trained with {\em Down@Train} is expected to be able to detect small-size instances, we input the original test images to the models without any scaling. Table \ref{tbl:DAT} shows the results. We can observe that {\em Down@Train} (i.e., AP$_{S}$) does improve the detection scores for small-size instances compared with the model trained without any small-size instance (i.e., AP$_{S^\dagger}$) but they are still lower than the original scores (i.e., AP$_{S^*}$). 

\begin{table}[t]
\footnotesize\centering
\caption{
Results of {\em Down@Train}. The training images are downscaled by a factor of $\beta$ and mixed with the original images. ${\rm AP}_{S^*}$ shows the small-instance detection score of {\em Baseline} and ${\rm AP}_{S^\dagger}$ shows that of the detectors trained only on medium- and large-size instances using unscaled images (i.e., {\em  Up@Test-vanilla} at $\alpha=1.0$.) ${\rm AP}_S$, ${\rm AP}_M$, and ${\rm AP}_L$ are the scores for small/medium/large instances. 
}
\vskip 0.1in
\label{tbl:DAT}
\scalebox{0.89}{
\begin{tabular}{cc|c|c|ccc|c}\hline
Detector & $\beta$ & ${\rm AP}_{S^*}$ & ${\rm AP}_{S^\dagger}$  & ${\rm AP}_S$ & ${\rm AP}_M$ & ${\rm AP}_L$ & ${\rm AP}$ \\
\hline
            & 1/2 &      &     & 7.3 & 36.5 & 45.5 & 28.1 \\
Faster RCNN & 1/3 & 11.9 & 5.2 & 6.6 & 35.6 & 45.1 & 27.5 \\ 
            & 1/4 &      &     & 6.5 & 35.2 & 45.1 & 27.2 \\ \hline
            & 1/2 &      &     & 14.5 & 47.2 & 53.7 & 34.7 \\
FCOS        & 1/3 & 22.3 & 7.7 & 13.6 & 46.9 & 52.3 & 34.3 \\
            & 1/4 &      &     & 14.4 & 45.7 & 53.0 & 33.8 \\
\hline
\end{tabular}
}
\end{table}

\subsection{Summary of the Results}
\label{sec:discussion}

\begin{table}[tb]\footnotesize\centering
\caption{Summary of experimental results.}
\vskip 0.1in
\label{tbl:summary}
\scalebox{0.9}{
\begin{tabular}{c|ccc} \hline
    \multicolumn{1}{c|}{} & \multicolumn{3}{c}{Test on original small-size instance} 
    \\ \cline{2-4}
    & {\em Baseline} & {\em Up@Test}($\gamma=3$) & {\em Down@Train}($\beta=1/3$)
    \\ \hline\hline
    No domain gap & \checkmark
    \\
    W/o small anno.
    & & \checkmark & \checkmark
    \\
    Input upscaling & & \checkmark 
    \\ \hline
    Faster RCNN & 11.9 & 13.7($\alpha=2.0$) &  6.6
    \\ \hline
    FCOS & 22.3 & 23.6($\alpha=2.2$) & 13.6 
    \\ \hline
\end{tabular}
}
\end{table}

Table \ref{tbl:summary} shows the summary of the  results we have seen so far. We have evaluated three methods (i.e., {\em Baseline}, {\em Up@Test}, and {\em Down@Train}). The three methods can be characterized by the following three factors: the presence of a domain gap between train and test inputs due to image scaling, the availability of small-size object annotation, and input upscaling (i.e., upscaling the input image at test time).

We can make the following observations:

\medskip\noindent 
i) {\em Detectors can learn to detect small-size instances when there is no real small-size instance annotation.} To achieve the same level of accuracy, it suffices to only use test-time upscaling. This is confirmed by the higher performance of {\em Up@Test-scale} than {\em Baseline}.

\medskip\noindent 
ii) Downscaling object instances at training time does not perform well. This is arguably due to the domain gap between the original and the downscaled annotations for small-size instances. It should be caused by the factors other than image resolution, i.e., differences in the context and perspective of objects.  

\medskip\noindent
iii) Upscaling an input image at test time is so effective that it can recover the loss caused by the domain gap.

\subsection{Distilled Single-path Model}

As shown above, {\em Up@Test} achieves comparable or even better performance than {\em Baseline}. However, it feeds an input image into a detector twice with different scaling, and it may not be fair to conclude {\em Up@Test} matches {\em Baseline}. Thus, we consider distillation of an {\em Up@Test} model to a single-path baseline detector. 

\begin{figure}[tb]
\centering
\includegraphics[width=0.9\linewidth,bb=0 0 2918 2300]{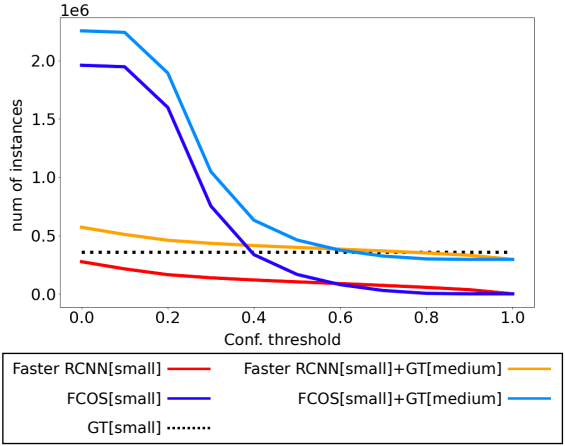}
\caption{
The number of instances that are detected for all the COCO training images by {\em Up@Test-scale} with a varying threshold of confidence (i.e., the curves in red and blue). The dotted line indicates the number of original small instances. When using downscaled medium-size instances as small-size instances, their combined population is similar to the original ones. 
}
\label{fig:num_distilled}
\end{figure}

We perform the distillation in the following way. First, we apply an {\em Up@Test} model to all the images of the training set of the COCO dataset, obtaining detection results. We then choose detection results with a confidence score larger than a specified threshold and also a bounding box having a size in the range of small object instances. However, when we set the threshold larger than 0.5, the population of the instances tends to be way smaller than that of the original small-size instances, as shown in Fig.~\ref{fig:num_distilled}. This causes a severe imbalance in the number of instances between different instance sizes, leading to mediocre performance. Thus, we additionally use {\em Down@Train} to generate instances by downscaling medium and large-size instance annotations. Merging all the annotations, i.e., pseudo and downscaled small-size instances and the original medium and large-size instances, we train a student model with them. We use the same detector for a student model. 

Table \ref{tbl:dist-uptest+downtrain} shows the results. We used {\em Up@Test-scale} with $\alpha=2.0$ for Faster RCNN and $2.2$ for FCOS and $\gamma=3.0$ for the both. It is seen that the best models for Faster RCNN and FCOS achieve 11.0 and 22.4 in AP$_S$, respectively. They are comparable to their base detectors trained with the full training data. It is noteworthy that the distilled detectors do not lose the accuracy for medium and large-size instances (i.e., AP$_M$ and AP$_L$). These will reinforce our claim that the annotation of small-size instances is less important than we thought.

\begin{table}[t]
\footnotesize\centering
\caption{
Results of the distilled single-path model. `Conf. thresh.' is the threshold for pseudo labels used for training.
}
\vskip 0.1in
\label{tbl:dist-uptest+downtrain}
\scalebox{0.93}{
\begin{tabular}{c|c|c|c|c|c}\hline
Detector & Method & Conf. thresh. & ${\rm AP}_S$ & ${\rm AP}_M$ & ${\rm AP}_L$ \\ \hline
\multirow{4}{*}{Faster RCNN} & Baseline & - & 11.9 & 34.9 & 46.1 \\ \cline{2-6}
                             & & 0.3 & 10.1 & 34.4 & 44.5 \\
                             & Distilled & 0.7 & 10.3 & 35.0 & 45.3 \\
                             & & 0.9 & 11.0 & 35.6 & 45.0 \\ \hline
\multirow{4}{*}{FCOS} & Baseline & - & 22.3 & 43.7 & 52.5 \\ \cline{2-6}
                             & & 0.3 & 13.6 & 44.3 & 51.0 \\
                             & Distilled & 0.7 & 22.4 & 46.5 & 52.7 \\ 
                             & & 0.9 & 15.2 & 46.7 & 52.5 \\ \hline
\end{tabular}
}
\end{table}

\section{Summary and Conclusion}

We have reconsidered the annotation of training data for object detection. Considering that it is more costly than larger-size instances, we questioned if annotating small-size object instances is worth its cost. To study this question, we pose it differently: can we detect small-size object instances with a detector trained using training data free of small-size instance annotation?

To answer this question, we considered two methods and experimentally evaluated their performance using the COCO dataset. One is to upscale images at test time before inputting them to a detector. The other is to downscale images at training time to create synthetic small-size instances and train a detector using them. We expect the first approach to suffer from a domain gap between training and test inputs since the upscaled small-size instances will have low image resolution. This gap could affect the detector's performance. We considered an extended method to cope with this, which trains a detector using blurred versions of medium and large-size instances along with their originals.

The experimental results show that the first method yields at least comparable performance to the baseline detector trained using the entire training data. This is an answer to the above question: we can detect small-size instances with a detector trained using small-size instance free data. The method needs to apply the detector to the same image twice with different scaling to detect all size instances. We have shown that we can distill the model into a single-path detector which we can use in an equal manner to the baseline detector. These results point to the necessity of rethinking the annotation of small-size objects for object detection.

%%%%%%%%%%%%%%%%%%%%%%%%%%%%%%%%%%%%%%%%%%%%%%%%%%%%%%%%%%%%
{\small
\bibliographystyle{ieee_fullname}
\bibliography{egbib}

\begin{thebibliography}{10}\itemsep=-1pt

\bibitem{CascadeRCNN}
Z. Cai and N. Vasconcelos.
\newblock {Cascade R-CNN: Delving Into High Quality Object Detection}.
\newblock In {\em Proc. CVPR}, 2018.

\bibitem{R_FCN}
J. Dai, Y. Li, K. He, and J. Sun.
\newblock {R-FCN: Object Detection via Region-based Fully Convolutional Networks}.
\newblock In {\em Proc. NeurIPS}. 2016.

\bibitem{CenterNet}
K. Duan, S. Bai, L. Xie, H. Qi, Q. Huang, and Q. Tian.
\newblock {CenterNet: Keypoint Triplets for Object Detection}.
\newblock In {\em Proc. ICCV}, 2019.

\bibitem{VOC}
M. Everingham, L. V.~Gool, C.~K.~I. Williams, J. Winn, and A. Zisserman.
\newblock {The Pascal Visual Object Classes (VOC) Challenge}.
\newblock {\em IJCV}, 2010.

\bibitem{DSSD}
C.~Y. Fu, W. Liu, A. Ranga, A. Tyagi, and A.~C. Berg.
\newblock {{DSSD} : Deconvolutional Single Shot Detector}.
\newblock {\em arXiv:1701.06659}, 2017.

\bibitem{NoteRCNN}
J. Gao, J. Wang, S. Dai, L.-J. Li, and R. Nevatia.
\newblock {NOTE-RCNN: NOise Tolerant Ensemble RCNN for Semi-supervised Object Detection}.
\newblock In {\em Proc. ICCV}, 2019.

\bibitem{KITTI}
A. Geiger, P. Lenz, and R. Urtasun.
\newblock {Are We Ready for Autonomous Driving? The KITTI Vision Benchmark Suite}.
\newblock In {\em Proc. CVPR}, 2012.

\bibitem{coteaching}
B. Han, Q. Yao, X. Yu, G. Niu, M. Xu, W. Hu, I. Tsang, and M. Sugiyama.
\newblock {Co-teaching: Robust Training of Deep Neural Networks with Extremely Noisy Labels}.
\newblock In {\em Proc. NeurIPS}, 2018.

\bibitem{MaskRCNN}
K. He, G. Gkioxari, P. Dollar, and R. Girshick.
\newblock {Mask R-CNN}.
\newblock In {\em Proc. ICCV}, 2017.

\bibitem{ResNet}
K. He, X. Zhang, S. Ren, and J. Sun.
\newblock {Deep Residual Learning for Image Recognition}.
\newblock In {\em Proc. CVPR}, 2016.

\bibitem{ConsistencySSOD}
J. Jeong, S. Lee, J. Kim, and N. Kwak.
\newblock {Consistency-based Semi-supervised Learning for Object detection}.
\newblock In {\em Proc. NeurIPS}, 2019.

\bibitem{mentornet}
L. Jiang, Z. Zhou, T. Leung, L.-J. Li, and L. Fei-Fei.
\newblock {MentorNet: Learning Data-Driven Curriculum for Very Deep Neural Networks on Corrupted Labels}.
\newblock In {\em Proc. ICML}, 2018.

\bibitem{FoveaBox}
T. Kong, F. Sun, H. Liu, Y. Jiang, and J. Shi.
\newblock {FoveaBox: Beyond Anchor-based Object Detector}.
\newblock {\em arXiv:1904.03797}, 2019.

\bibitem{LearnDialog}
K. Konyushkova, J. Uijlings, C.~H. Lampert, and V. Ferrari.
\newblock {Learning Intelligent Dialogs for Bounding Box Annotation}.
\newblock In {\em Proc. CVPR}, 2018.

\bibitem{ObjectAware}
S. Kosugi, T. Yamasaki, and K. Aizawa.
\newblock {Object-aware Instance Labeling for Weakly Supervised Object Detection}.
\newblock In {\em Proc. ICCV}, 2019.

\bibitem{OpenImages}
A. Kuznetsova, H. Rom, N. Alldrin, J. Uijlings, I. Krasin, J. Pont-Tuset, S. Kamali, S. Popov, M. Malloci, A. Kolesnikov, T. Duerig, and V. Ferrari.
\newblock {The Open Images Dataset V4: Unified Image Classification, Object Detection, and Visual Relationship Detection at Scale}.
\newblock {\em IJCV}, 2020.

\bibitem{CornerNet}
H. Law and J. Deng.
\newblock {CornerNet: Detecting Objects as Paired Keypoints}.
\newblock In {\em Proc. ECCV}, 2018.

\bibitem{NR_OD}
J. Li, C. Xiong, R. Socher, and S. Hoi.
\newblock {Towards Noise-resistant Object Detection with Noisy Annotations}.
\newblock {\em arXiv:2003.01285}, 2020.

\bibitem{MSCOCO}
T.{-}Y. Lin, M. Maire, S.~J. Belongie, L.~D. Bourdev, R.~B. Girshick, J. Hays, P. Perona, D. Ramanan, P. Doll{\'{a}}r, and C.~L. Zitnick.
\newblock {Microsoft {COCO:} Common Objects in Context}.
\newblock In {\em Proc. ECCV}, 2014.

\bibitem{FPN}
T.~Y. Lin, P. Dollar, R. Girshick, K. He, B. Hariharan, and S. Belongie.
\newblock {Feature Pyramid Networks for Object Detection}.
\newblock In {\em Proc. CVPR}, 2017.

\bibitem{RetinaNet}
T.~Y. Lin, P. Goyal, R. Girshick, K. He, and P. Dollar.
\newblock {Focal Loss for Dense Object Detection}.
\newblock In {\em Proc. ICCV}, 2017.

\bibitem{PANet}
S. Liu, L. Qi, H. Qin, J. Shi, and J. Jia.
\newblock {Path Aggregation Network for Instance Segmentation}.
\newblock In {\em Proc. CVPR}, 2018.

\bibitem{SSD}
W. Liu, D. Anguelov, D. Erhan, C. Szegedy, S. Reed, C. Fu, and A.~C. Berg.
\newblock {SSD: Single Shot MultiBox Detector}.
\newblock In {\em Proc. ECCV}, 2016.

\bibitem{GridRCNN}
X. Lu, B. Li, Y. Yue, Q. Li, and J. Yan.
\newblock {Grid R-CNN}.
\newblock In {\em Proc. CVPR}, 2019.

\bibitem{Decoupling}
E. Malach and S. Shalev-Shwartz.
\newblock {Decoupling "When to Update" from "How to Update"}.
\newblock In {\em Proc. NeurIPS}, 2017.

\bibitem{AverageDelay}
H. Mao, X. Yang, and W.~J. Dally.
\newblock {A Delay Metric for Video Object Detection: What Average Precision Fails to Tell}.
\newblock In {\em Proc. ICCV}, 2019.

\bibitem{GS_Face}
X. Ming, F. Wei, T. Zhang, D. Chen, and F. Wen.
\newblock {Group Sampling for Scale Invariant Face Detection}.
\newblock In {\em Proc. CVPR}, 2019.

\bibitem{LRP_Det}
K. Oksuz, B.~C. Cam, E. Akbas, and S. Kalkan.
\newblock {Localization Recall Precision (LRP): A New Performance Metric for Object Detection}.
\newblock In {\em Proc. ECCV}, 2018.

\bibitem{ExtremeClick}
D.~P. Papadopoulos, J.~R.~R. Uijlings, F. Keller, and V. Ferrari.
\newblock {Extreme Clicking for Efficient Object Annotation}.
\newblock In {\em Proc. ICCV}, 2017.

\bibitem{ClickSV}
D.~P. Papadopoulos, J.~R.~R. Uijlings, F. Keller, and V. Ferrari.
\newblock {Training Object Class Detectors with Click Supervision}.
\newblock In {\em Proc. CVPR}, 2017.

\bibitem{yolov2}
J. Redmon and A. Farhadi.
\newblock {YOLO9000: Better, Faster, Stronger}.
\newblock In {\em Proc. CVPR}, 2017.

\bibitem{yolov3}
J. Redmon and A. Farhadi.
\newblock {YOLOv3: An Incremental Improvement}.
\newblock {\em arXiv:1804.02767}, 2018.

\bibitem{LearnReweight}
M. Ren, W. Zeng, B. Yang, and R. Urtasun.
\newblock {Learning to Reweight Examples for Robust Deep Learning}.
\newblock In {\em Proc. ICML}, 2018.

\bibitem{FasterRCNN}
S. Ren, K. He, R. Girshick, and J. Sun.
\newblock {Faster R-CNN: Towards Real-time Object Detection with Region Proposal Networks}.
\newblock In {\em Proc. NeurIPS}, 2015.

\bibitem{AppearanceTrans}
M. Rochan and Y. Wang.
\newblock {Weakly Supervised Localization of Novel Objects Using Appearance Transfer}.
\newblock In {\em Proc. CVPR}, 2015.

\bibitem{ReduceNoise}
N. Samet, S. Hicsonmez, and E. Akbas.
\newblock {Reducing Label Noise in Anchor-free Object Detection}.
\newblock In {\em Proc. BMVC}, 2020.

\bibitem{SNIP}
B. Singh and L.~S. Davis.
\newblock {An Analysis of Scale Invariance in Object Detection ^^c2^^ad SNIP}.
\newblock In {\em Proc. CVPR}, 2018.

\bibitem{SNIPER}
B. Singh, M. Najibi, and Larry~S Davis.
\newblock {SNIPER: Efficient Multi-scale Training}.
\newblock In {\em Proc. NeurIPS}. 2018.

\bibitem{EFDet}
M. Tan, R. Pang, and Q.~V. Le.
\newblock {EfficientDet: Scalable and Efficient Object Detection}.
\newblock In {\em Proc. CVPR}, 2020.

\bibitem{JointOptim}
D. Tanaka, D. Ikami, T. Yamasaki, and K. Aizawa.
\newblock {Joint Optimization Framework for Learning with Noisy Labels}.
\newblock In {\em Proc. CVPR}, 2018.

\bibitem{LS_SSOD}
Y. Tang, J. Wang, B. Gao, E. Dellandrea, R. Gaizauskas, and L. Chen.
\newblock {Large Scale Semi-supervised Object Detection Using Visual and Semantic Knowledge Transfer}.
\newblock In {\em Proc. CVPR}, 2016.

\bibitem{FCOS}
Z. Tian, C. Shen, H. Chen, and T. He.
\newblock {FCOS: Fully Convolutional One-stage Object Detection}.
\newblock In {\em Proc. ICCV}, 2019.

\bibitem{FixResolution}
H. Touvron, A. Vedaldi, M. Douze, and H. Jegou.
\newblock {Fixing the Train-test Resolution Discrepancy}.
\newblock In {\em Proc. NeurIPS}. 2019.

\bibitem{RevisitKnowTrans}
J. Uijlings, S. Popov, and V. Ferrari.
\newblock {Revisiting Knowledge Transfer for Training Object Class Detectors}.
\newblock In {\em Proc. CVPR}, 2018.

\bibitem{RepPoints}
Z. Yang, S. Liu, H. Hu, L. Wang, and S. Lin.
\newblock {RepPoints: Point Set Representation for Object Detection}.
\newblock In {\em Proc. ICCV}, 2019.

\bibitem{ProbNoiseCorrection}
K. Yi and J. Wu.
\newblock {Probabilistic End-to-end Noise Correction for Learning with Noisy Labels}.
\newblock In {\em Proc. CVPR}, 2019.

\bibitem{RefineDet}
S. Zhang, L. Wen, X. Bian, Z. Lei, and S.~Z. Li.
\newblock {Single-shot Refinement Neural Network for Object Detection}.
\newblock In {\em Proc. CVPR}, 2018.

\bibitem{LatentV}
X. Zhang, Y. Yang, and J. Feng.
\newblock {Learning to Localize Objects with Noisy Labeled Instances}.
\newblock In {\em Proc. AAAI}, 2019.

\bibitem{M2Det}
Q. Zhao, T. Sheng, Y. Wang, Z. Tang, Y. Chen, L. Cai, and H. Ling.
\newblock {M2det: A Single-shot Object Detector Based on Multi-level Feature Pyramid Network}.
\newblock In {\em Proc. AAAI}, 2019.

\bibitem{ObjectsAsPoints}
X. Zhou, D. Wang, and P. Kr{\"{a}}henb{\"{u}}hl.
\newblock {Objects as Points}.
\newblock {\em arXiv:1904.07850}, 2019.

\bibitem{ExtremeNet}
X. Zhou, J. Zhuo, and P. Krahenbuhl.
\newblock {Bottom-up Object Detection by Grouping Extreme and Center Points}.
\newblock In {\em Proc. CVPR}, 2019.

\bibitem{FSFA}
C. Zhu, Y. He, and M. Savvides.
\newblock {Feature Selective Anchor-free Module for Single-shot Object Detection}.
\newblock In {\em Proc. CVPR}, 2019.

\end{thebibliography}
}
%%%%%%%%%%%%%%%%%%%%%%%%%%%%%%%%%%%%%%%%%%%%%%%%%%%%%%%%%%%%
\clearpage
\appendix

\section*{\Large{Appendix}}

\section{Examples of Detection Results}

As is reported in the main paper, {\em Up@Test-scale} is comparable or slightly better than {\em Baseline} in terms of small-size instance detection accuracy in the evaluation using original small-size instances. It is much better than {\em Baseline} in the evaluation using synthetic small-size instances (i.e., downscaled medium-size instances). {\em DAT} performs better than {\em Baseline} in the latter evaluation. We show here several detection results for these three method-evaluation pairs showing their success and failure cases. 

\subsection{{\bf\em Baseline} vs. {\bf\em Up@Test-scale} in Evaluation with Original Small Instances}

\begin{figure*}[tb]
\centering
\includegraphics[width=0.8\linewidth,bb=0 0 2000 2828]{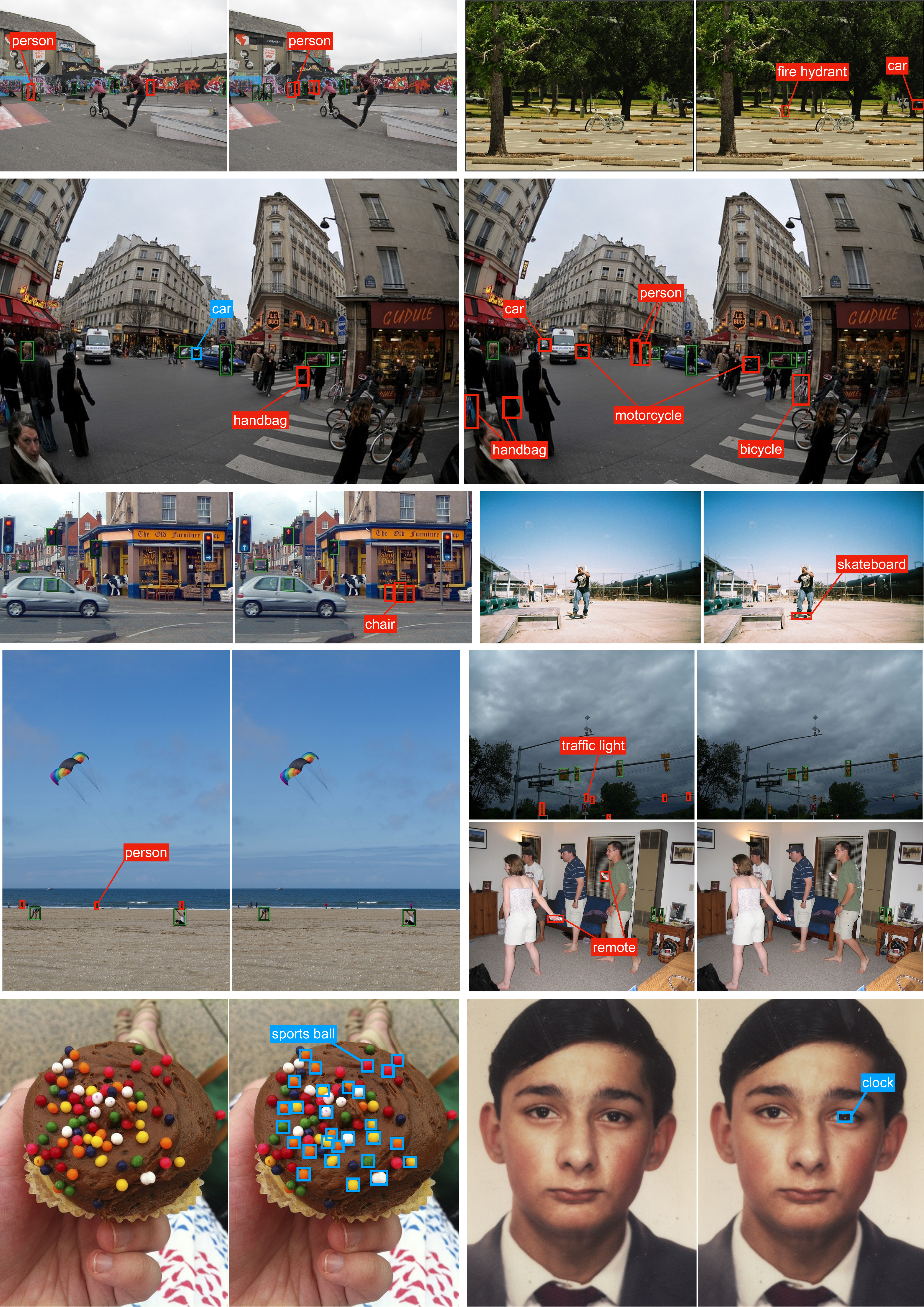}
\caption{Success and failure cases of small-size instance detection evaluated by original test small-size instances by {\em Baseline} (left of each image pair) and {\em Up@Test-scale} (right). The parameters of the latter are $\gamma=3$ and $\alpha=2.2$. Green BBs are the true positives detected by both methods; the red BBs are true positives detected by only one of the two; the light blue BBs are false positives.}
\label{fig:original_uat}
\end{figure*}

Figure \ref{fig:original_uat} shows several results of  {\em Baseline} and {\em Up@Test-scale} ($\gamma=3$ and $\alpha=2.2$) evaluated with original small-size instances. We use FCOS and set the confidence threshold for BBs to $0.5$. Each image contain BBs in three colors; the green BBs are the true positives detected by both two methods; the red BBs are true positives detected by only one of the two; the light blue BBs are false positives.  

We can observe that {\em Up@Test-scale} can detect several small instances that {\em Baseline} cannot detect in some images, and it fails to detect some instances that {\em Baseline} detects correctly in other images. We think these differences arise from a combination of several reasons. One is the distributional differences of instance sizes of objects. Specifically, for objects that have a sufficient number of medium- and large-size instances in the training data (and maybe a smaller number of small-size instances), {\em Up@Test-scale} may detect small-size instances more accurately than {\em Baseline}. Examples are ``fire hydrant'' and ``skateboard'' in Fig.~\ref{fig:original_uat}. The opposite may be true; for objects that have only small-size instances in the training data (e.g., ``traffic light'' and ``remote''), {\em Baseline} performs better. 

{\em Up@Test-scale} tends to misdetect extremely small-size instances, such as ``person'' located far away from the camera; we think this is either because {\em Up@Test-scale} cannot use the objects' context properly or because {\em Baseline} may rather be overfit to them, as they seem too small to identify. Furthermore, {\em Up@Test-scale} sometimes yield many false-positives that {\em Baseline} does not. We conjecture this is also because {\em Up@Test-scale} cannot properly use their context, e.g., ``sports ball'' on a cup cake and ``clock'' in an eye. These imply there is still room for improvement in the upscale-at-test-time approach. We will examine these in a future study. 

\end{document}